\documentclass[runningheads]{llncs}
\usepackage{graphicx}
\usepackage{subfig}
\usepackage{amsmath,amssymb} 
\usepackage{bbm}
\usepackage{color}

\usepackage{threeparttable}
\usepackage{url}
\usepackage{multirow}
\usepackage{makecell}

\usepackage[pagebackref=true,breaklinks=true,letterpaper=true,colorlinks,bookmarks=false]{hyperref}

\DeclareMathOperator*{\argminA}{arg\,min}
\DeclareMathOperator*{\argmaxA}{arg\,max}

\newcommand{\thickhline}{%
    \noalign {\ifnum 0=`}\fi \hrule height 1pt
    \futurelet \reserved@a \@xhline
}
\newcolumntype{"}{@{\hskip\tabcolsep\vrule width 1pt\hskip\tabcolsep}}
\begin{document}
\pagestyle{headings}
\mainmatter

\def\ACCV20SubNumber{219}  

\title{MIX'EM: Unsupervised Image Classification using a Mixture of Embeddings} 
\titlerunning{MIX'EM: Unsupervised Image Classification using a Mixture of Embeddings}
%
\author{Ali Varamesh\inst{1} \and
Tinne Tuytelaars\inst{1}}
\authorrunning{A. Varamesh et al.}
%
\institute{ESAT-PSI, KU Leuven\\
\email{\{ali.varamesh,tinne.tuytelaars\}@esat.kuleuven.be}}

\maketitle
\begin{abstract}
We present MIX'EM, a novel solution for unsupervised image classification. MIX'EM generates representations that by themselves are sufficient to drive a general-purpose clustering algorithm to deliver high-quality classification. This is accomplished by building a mixture of embeddings module into a contrastive visual representation learning framework in order to disentangle representations at the category level. It first generates a set of embedding and mixing coefficients from a given visual representation, and then combines them into a single embedding. We introduce three techniques to successfully train MIX'EM and avoid degenerate solutions; (i) diversify mixture components by maximizing entropy, (ii) minimize instance conditioned component entropy to enforce a clustered embedding space, and (iii) use an associative embedding loss to enforce semantic separability. By applying (i) and (ii), semantic categories emerge through the mixture coefficients, making it possible to apply (iii). Subsequently, we run K-means on the representations to acquire semantic classification. We conduct extensive experiments and analyses on STL10, CIFAR10, and CIFAR100-20 datasets, achieving state-of-the-art classification accuracy of 78\%, 82\%, and 44\%, respectively. To achieve robust and high accuracy, it is essential to use the mixture components to initialize K-means. Finally, we report competitive baselines (70\% on STL10) obtained by applying K-means to the "normalized" representations learned using the contrastive loss.




\end{abstract}

\section{Introduction}

In the span of a few years, supervised image classification has made remarkable progress and even surpassed humans on specific recognition tasks \cite{he2016deep}. Its success depends on few important factors, namely, stochastic gradient descent to optimize millions of parameters, GPUs to accelerate high-dimensional matrix computations, and access to vast amounts of manually annotated data \cite{krizhevsky2012imagenet}. Although a particular optimization method or high-performance hardware is not theoretically essential for supervised learning methods, labeled data is. In fact, access to large-scale labeled data is vital if we want to get the top performance \cite{sun2017revisiting,he2019rethinking}. Hence, one of the current major challenges in modern computer vision is being able to do unsupervised visual recognition. That means eliminating the costly and not always feasible process of manual labeling \cite{lin2014microsoft,deng2009imagenet}. 
In this context, visual representation learning has recently demonstrated great success in discarding manual labels by relying on self-supervision \cite{bachman2019learning,gidaris2018unsupervised,wu2018unsupervised,oord2018representation,chen2020simple,he2020momentum,tian2019contrastive}. We believe self-supervised representation learning has paved the way for unsupervised recognition.  


\begin{figure}[t]
\centering
\includegraphics[width=120mm]{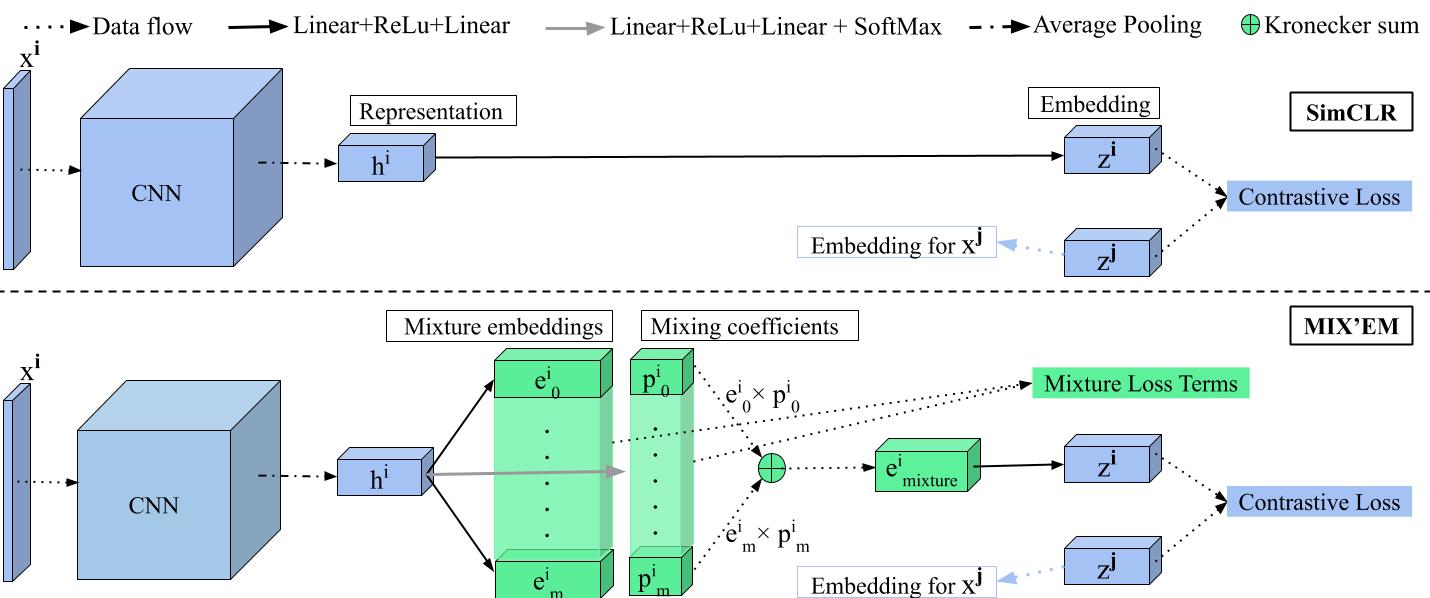} 
\caption{Our proposed architecture with a mixture of embeddings module (bottom row), compared to 
the contrastive representation learning framework (top row). Components in green are devised in MIX'EM. 
}
\label{fig:schematics}
\end{figure}

In self-supervised visual representation learning, a pretext (auxiliary) task provides a supervision signal (without manual labels) for training a representation encoder. One particularly successful pretext task is to treat each image instance in a dataset as a unique class and use them as supervision labels for training a classifier using the cross-entropy objective \cite{wu2018unsupervised}. However, it is computationally prohibitive to implement at a scale of millions of images. In practice, it is simplified such that given a mini-batch containing transformations of different images, transformations of a particular image should be classified as the same \cite{he2020momentum,chen2020simple}. By training a linear classifier on representations generated by such an encoder, we can achieve high accuracy, close to that of an end-to-end trained fully supervised ImageNet classifier: 76.5\% compared to  78.3\% in terms of top-1 accuracy, respectively. \cite{chen2020simple}. They have even outperformed supervised representations on some variants of object detection and segmentation \cite{he2020momentum}. 


The fact that a linear layer, with very limited discriminative power \cite{asano2019critical}, can deliver such high accuracy on the complex ImageNet classification task signals presence of powerful semantic clues in the representations. Hence, we hypothesize that by just knowing the expected number of classes, an off-the-shelf clustering method must be able to deliver high accuracy clustering similarly. However, our experiments show that K-means trained on "normalized" representations generated by the recent SimCLR method \cite{chen2020simple} achieves clustering accuracy of 70\% on STL10 compared to top-1 accuracy of 87\% by a supervised linear classifier. Therefore there is significant room for improvement.

Our goal in this work is to impose semantic structure on the self-supervised representations to boost clustering accuracy. In other words, we want to generate representations that are already highly clustered or disentangled. For this purpose, we build a mixture of embeddings module into the contrastive visual representation learning framework \cite{chen2020simple}, as illustrated in figure \ref{fig:schematics}. The mixture components \cite{mclachlan1988mixture,jordan1994hierarchical,bishop1994mixture} are expected to specialise on embedding different semantic categories. For a given sample, each component should generate an embedding and predict how much it contributes to the combined final embedding. We have designed MIX'EM essentially by taking inspiration from a few recent works \cite{greff2019multi,chen2019unsupervised,li2019generating,lee2015m,ye2018occlusion,makansi2019overcoming,varamesh2020mixture} showing that mixture models can divide their input-output space in a meaningful manner without being directly supervised. MIX'EM takes advantage of the contrastive visual learning framework for guiding training a mixture model without interfering with its mechanisms (see table \ref{table:linear}).



\begin{figure}[t]
\centering
\includegraphics[width=120mm]{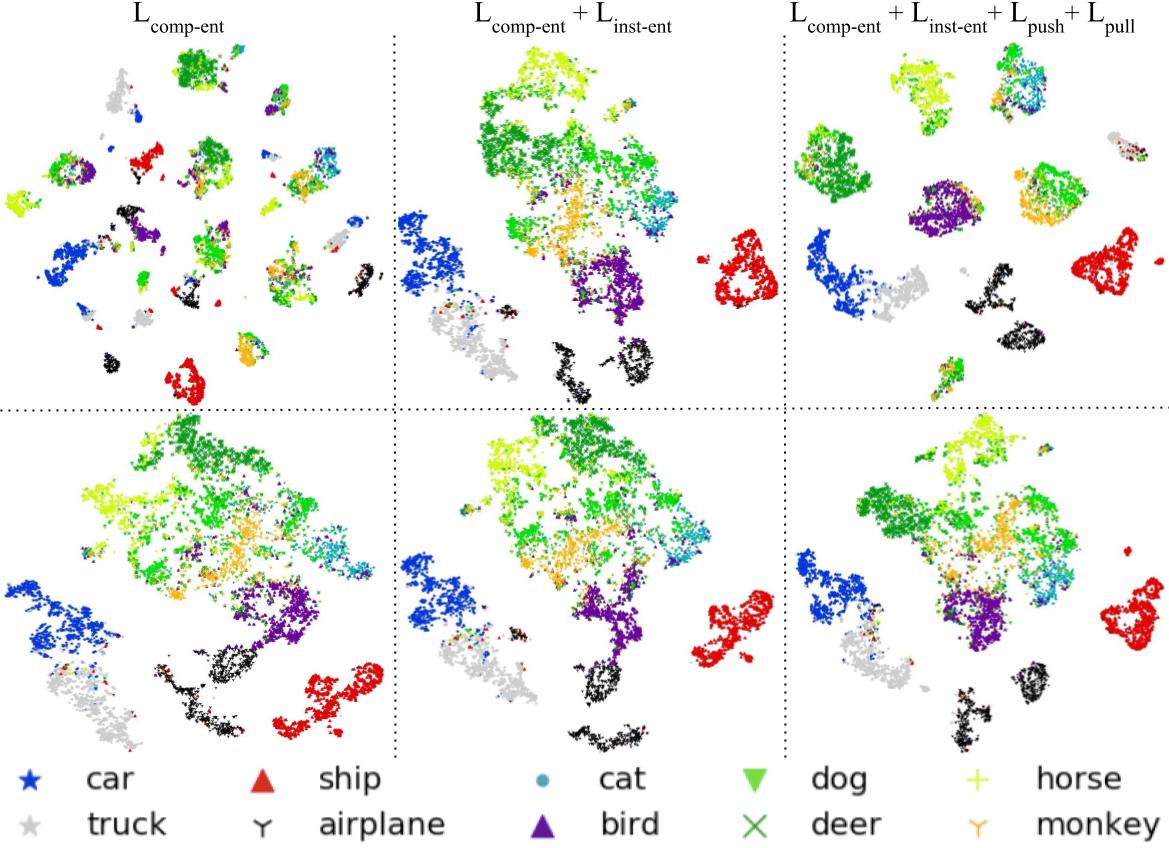} 
\caption{tSNE visualization of embeddings learned by MIX'EM on STL10. Top row: embeddings from the dominant component ($e_{m'}^i$ s.t. $m'=\argmaxA_{m'}(p_{m'}^i)$\ ); Bottom row: the final mixture embedding ($z^i$). By adding our loss terms (explained in \ref{loss_terms}), embedding space gets increasingly more disentangled at category level. Samples are marked and color coded based on their ground truth label.  
}
\label{fig:tsne_fcmax}
\end{figure}

In addition to the contrastive representation learning loss, we introduce three key techniques to successfully train MIX'EM end-to-end. A naive attempt to train using only the contrastive loss would quickly converge to a degenerate solution that assigns all samples to a single component, bypassing all other paths. To avoid this issue and achieve high accuracy, we (i) maximize the entropy of coefficient distribution to diversify the mixture components; (ii) minimize the entropy of components conditioned on the input to enforce separation of the embedding space; and enabled by (i) and (ii), (iii) we use an associative embedding loss \cite{newell2017pixels,newell2017associative} to directly enforce semantic coherency inter/intra mixture components. Figure \ref{fig:tsne_fcmax} presents visualizations of the embeddings when gradually plugging in each of the loss terms. The resulting representations significantly boost K-means' performance up to 78\% on the STL10 dataset, without interfering with the contrastive learning process. We summarise our contributions as follows:

\begin{itemize}
\item We propose MIX'EM, a solution for unsupervised image classification using a mixture of embedding module. MIX'EM disentangles visual representations semantically at the category level, such that an off-the-shelf clustering method can be applied to acquire robust image classification.
\item We introduce three techniques to successfully train MIX'EM in an unsupervised manner and avoid degenerate solutions.
\item We introduce a technique to initialize K-means algorithm using the mixture components in MIX'EM and achieve significantly higher accuracy. This eliminates the need to run K-means with multiple random initializations.

\end{itemize}

\section{Related Work}

Our work relates to a few different lines of research. It is the most related to the self-supervised representation learning, as our goal in the first place is to train a better representation encoder without using manually labeled data. However, beyond that, we want the representations to be highly structured such that reliable semantic clustering is possible using an off-the-shelf clustering method. We develop our idea on the recent research \cite{chen2020simple,chen2020improved} which empirically proves using noise contrastive loss \cite{gutmann2010noise} and heavy augmentation for visual representation learning outperforms other popular approaches, including mutual information maximization \cite{hjelm2018learning,bachman2019learning}, generative models \cite{donahue2016adversarial,dumoulin2016adversarially}, image rotation prediction \cite{gidaris2018unsupervised}, predicting patch position \cite{doersch2015unsupervised}, clustering \cite{caron2018deep}, solving jigsaw puzzles \cite{noroozi2016unsupervised}, and image colorization \cite{zhang2016colorful}.
In this work, we also advocate using contrastive loss for self-supervised representation learning. However, we are particularly interested in enforcing category-level semantic disentanglement on the representation space.


To the best of our knowledge, this is the first work to set out to impose semantic structure on self-supervised visual representations learned using the contrastive loss. We show that the representations generated by MIX'EM result in high accuracy semantic clustering only by applying K-means to them. Existing works on unsupervised image classification using self-supervised representations \cite{van2020learning,huang2020deep,ji2019invariant} should benefit from adapting our proposed module, as it is an internal module that can be plugged-in without altering the output mechanism. 

There have been a few recent works with the same objective as ours, that is unsupervised image classification. IIC \cite{ji2019invariant} is the best known among them, which directly generates semantic clustering assignments using a deep network. Its loss function maximizes the mutual information between augmented versions of the same image based on the cluster assignment. The intuitive goal is to force the clustering to be decided based on invariant information across different views of an image. \cite{huang2020deep} proposes a max-margin clustering criterion for simultaneous clustering and representation learning, such that clustering confidence is the highest based on a defined confidence index. Finally, concurrent with our work, \cite{van2020learning} proposes a framework with multiple stages that relies on the k-nearest neighbors method to extract samples that are ideally from the same semantic categories based on their representations learned in a self-supervised fashion. They use the samples to train a clustering network, and then a classification network by treating clusters as pseudo labels. None of these works concern improving the representations directly in terms of category-level disentanglement in an unsupervised fashion. 

Our work also relates to clustering-based approaches for deep self-supervised representation learning \cite{caron2018deep,asano2019self,yang2016joint,yan2020clusterfit}. These models devise a branch in a deep network for clustering; that generates pseudo labels for training another branch of the network for a classification task. The training process either iterate between the two stages until it converges \cite{caron2018deep,asano2019self} or does it simultaneously \cite{zhan2020online}. Generating high-level semantic labels using clustering, however, is not the goal in this line of work. Instead, they combine clustering and classification in order to build a pretext task for representation learning. Often the best representations are achieved with over-clustering. For example, \cite{caron2018deep} achieves the best mAP on the Pascal VOC 2007 object detection task when the representations are learned using a 10000-way clustering stage. 

Finally, \cite{sanchez2019learning} is also related to our work, where representations are split into "shared" and "exclusive" parts. It maximizes mutual information for the shared and minimizes for the exclusive component across paired samples. However, they use supervision to pair images for training. Moreover, the work is not concerned with semantic clustering. Based on their problem formulation and results, the disentanglement focuses a foreground-background separation.

\section{Method}
In this section, first we review the contrastive learning framework as proposed in SimCLR \cite{chen2020simple} (the principles are similar to \cite{he2020momentum,wu2018unsupervised,tian2019contrastive}). Next, we show how to integrate a mixture of embeddings module in this framework.

\subsection{Contrastive learning of visual representation}

Contrastive learning of visual representations is built upon the intuition that different transformations of an image should have the same characteristics, which identifies them as bearing the same semantics. In practice, this means that given a dataset with images containing a single dominant object (like ImageNet or CIFAR10/100), an ideal encoder should map different augmentations of an image to a very compact neighborhood. This interpretation implies considering every image in a dataset as a distinct class and training a classification network using cross-entropy loss \cite{wu2018unsupervised}. However, having as many classes as the number of samples in a large dataset is not scalable. A streamlined version of this idea, SimCLR \cite{chen2020simple}, is based on doing instance classification within mini-batches of images. 



In SimCLR, the goal is to train an encoder to generate visual representations. We denote the encoder function with $f$ such that $h^i = f(x^i)$, where $x^i \in \textit{D}$ is an RGB image from the unlabeled dataset $\textit{D}$. Encoder $f$ is implemented using a deep convolutional neural network. Training then will proceed by contrasting representations $h^i$ in order to pull together similar images in the space. $h^i$ is the representation intended to be used by downstream tasks. However, Chen et al.~\cite{chen2020simple} show that, before computing the contrastive loss, applying a further non-linear $g$ layer to $h^i$ results in significant improvement. So in the following definitions, the contrastive loss will be computed on $z^i = g(h^i)$. 

At training, given a mini-batch of N images, $\{x^i\}_{i=1}^{N}$, every image is augmented twice using a sequence of random transformations to generate 2N samples $\{\hat{x}^j\}_{j=1}^{2N}$. Then, the similarity between every pair $u$ and $v$ of the 2N samples is computed using function $sim(u,v) = z_u^{T}.z_v/||z_u||||z_v||$. Next, counteractive loss for a positive pair (i.e. two augmentations of the same image $x_i$) is implemented in form of cross-entropy loss for a 2N-1 way classification task, where the logits are set to the pairwise similarities of a given view with its positive counterpart, and 2N-2 views from the remaining augmented samples. The contrastive loss $l_c(\hat{x}^{j_1},\hat{x}^{j_2})$ for a positive pair $\hat{x}^{j_1}$ and $\hat{x}^{j_2}$ (two views of the image $x^j$) is shown in the Equ.~(\ref{equ:ntxent}), where $\tau$ is a temperature parameter \cite{chen2020simple}. The contrastive loss is computed for both views of each of the N images. The total contrastive loss $L_{contrast}$ is shown in Equ.~(\ref{equ:ntxent_total}).

\begin{equation}
\begin{gathered}
        l_c(\hat{x}^{j_1},\hat{x}^{j_2}) = - \log{\frac{exp(sim(\hat{x}^{j_1},\hat{x}^{j_2})/\tau )}{ \sum_{k=1}^{2N}  \mathbbm{1}_{[k \neq {j_1}]} exp(sim(\hat{x}^{j_1},\hat{x}^k)/\tau )  }}\\
\end{gathered}
\label{equ:ntxent}
\end{equation}
\begin{equation}
\begin{gathered}
        L_{contrast}= \frac{1}{2N}\sum_{k=1}^{N} { l_c(\hat{x}^{k_1},\hat{x}^{k_2}) + l_c(\hat{x}^{k_2},\hat{x}^{k_1})}
\end{gathered}
\label{equ:ntxent_total}
\end{equation}

\subsection{Mixture Embedding}
\label{loss_terms}
SimCLR computes the contrastive loss after embedding the target representations ($h$) into another space ($z$) via the non-linear layer, $g$. In MIX'EM, we replace this layer with multiple parallel non-linear layers, each generating an embedding and a coefficient to determine how much the embedding contributes to the final embedding, $z^i$. Figure \ref{fig:schematics} depicts the architecture of MIX'EM and how it differs from the regular contrastive representation learning pipeline. Given input $x^i$, and representation $h^i = f(x^i)$ generated by the encoder, our model replaces $z^i = 
g(h^i)$ with the function $z^i = g(\psi(h^i))$, where the function $\psi$ is defined in Equ.~(\ref{equ:MIX'EM_module}). $M$ in Equ.~(\ref{equ:MIX'EM_module}) indicates the number of mixture components. $g_m$ is a non-linear layer similar to $g$ and specializes in generating embedding for samples that component $m$ is responsible for. Mixing coefficient $p_m^i$ indicates the prior probability of sample $x_i$ being generated by the component $m$. The coefficients $p^i$ for $x^i$ are computed from $h^i$ using a non-linear layer $g_p$ and softmax function.


\begin{equation}
        \psi(h^i) = \sum_{m=1}^{M} {p_m^i * g_m(h^i)})  \ \ \ \ \  s.t. \ \ \ \ \    p^i = softmax(g_p(h^i))
\label{equ:MIX'EM_module}
\end{equation}


With the mixture module, we expect the network to distribute input samples across components, as this should make the task easier \cite{mclachlan1988mixture,lee2015m}. Each mixture component should generate embeddings for certain semantic categories and guide the backpropagation process conditioned on the input. However,if we train MIX'EM only using the contrastive loss, it will quickly lead to a degenerate solution that assigns all samples to a single component. Therefore, we devise three loss terms to avoid such degenerate solutions and adequately train MIX'EM to meet our goals.

\subsubsection{Entropy maximization across components}
In a degenerate solution, the coefficients $p^i$ provide the lowest information from an information theory point of view; always, a particular component is equal to one. However, we want the model to be more diverse in the assignment of the components. We expect it to be dependent on the input image, not to ignore it. Given that we do not have any means to directly supervise the mixture module, we can instead maximize the entropy of the marginal distribution of mixtures $p$, which would take the highest value when all components are equally probable. As we will show in the experiments, this term indeed avoids the degenerate solution. Moreover, it will result in semantically meaningful components; that is, components will focus on different categories. We believe this is due to the simultaneous backpropagation of the contrastive loss, imposing minimal semantic regularization. In fact, without the contrastive loss, training would fail. The entropy maximization loss term is shown in Equ.~(\ref{equ:max_ebtropy_loss}), and is equal to the negative of entropy $\textit{H}(p)$.


\begin{equation}
\begin{gathered}
        L_{comp-ent} = - \textit{H}(p) = \sum{ p_m \log{p_m}}\ \ \ \ \  s.t. \ \ \ \ \ p_m = 1/N \sum_{i=0}^{N}{p_m^{i}} 
\end{gathered}
\label{equ:max_ebtropy_loss}
\end{equation}

\subsubsection{Conditional component entropy minimization}
Maximizing entropy of marginal $p$ diversifies the components. However, we would like to separate the representation space based on the most discriminative aspect of objects. For a given image, ideally, we want one of the mixture components to have close to full confidence so that it can be interpreted as an indicator of the true category. This, in turn, would mean reducing the entropy of the mixture components given an instance. We know that entropy would be minimized in practice if all probability mass is assigned to a single component. Therefore, given an image, we add a loss term that pushes the probability of the dominant (max) component to the highest value. Equ.~(\ref{equ:max_ebtropy_loss_cat}) shows the instance based entropy minimization loss term.

\begin{equation}
\begin{gathered}
        L_{inst-ent} = \sum_{i=0}^{N}{1- max\{p_m^{i}\}_{m=1}^{M} } 
        \\
\end{gathered}
\label{equ:max_ebtropy_loss_cat}
\end{equation}

\subsubsection{Associative embedding loss}
Both entropy-based loss terms above are principled techniques to guide the training. However, they do not directly take into account the semantics. Intuitively, samples' ideal assignment to the components should pick up on visual clues that minimize the distance between samples with the same dominant mixture component. At the same time, it should maximize the distance of samples with different dominant components. In a supervised setting, it is straightforward to implement a loss function like this given the true semantic labels; however, here we do not have access to such labels. The good news, however, is that just training MIX'EM with $L_{comp-ent}$ and $L_{inst-ent}$ would result in each component specializing in one category. In quantitative words, evaluating MIX'EM by treating the dominant component index as cluster label, on STL10, we get an accuracy of 73\% (row (3) of the fourth column in table \ref{table:incremental}.) 

Therefore, we introduce a third loss term to enforce semantic coherency by relying on the index of the dominant component as a pseudo-ground-truth label. This loss, called associative embedding, is inspired by the work of Newell et al.~\cite{newell2017pixels,newell2017associative} on scene graph generation and human pose estimation. Using the dominant component index as the class label, we want to pull the embeddings assigned to a component as close as possible to each other. We implement this by minimizing the distance of all embeddings by a component $m$ and the average embedding for the component on samples with $m$ as their dominant component (pull loss). Simultaneously, we wish to push the average embedding of different components away from each other (push loss). We implement this by directly maximizing the pairwise distance of the average embedding of components. Equations (\ref{equ:ae_means})-(\ref{equ:ae_push}) show formal specification of pull and push loss terms.  Note that $E_m$ is vital for the both losses, and we are able to compute it only by means of using the dominant components in MIX'EM.

\begin{equation}
\begin{gathered}
       \mu_m = \frac{1}{|E_m|}\sum_{i\in E_m}{e_m^i} \ \ \ \ \  s.t. \ \ \ \ \ E_m = \{i\ | \argminA_{m'}(p_{m'}^i)= m\}
\end{gathered}
\label{equ:ae_means}
\end{equation}

\begin{equation}
\begin{gathered}
       L_{pull}= \frac{1}{M} \sum_{m=1}^{M}{\sum_{i \in E_m}{|| e_m^i - \mu_m||_2} } \\
\end{gathered}
\label{equ:ae_pull}
\end{equation}

\begin{equation}
\begin{gathered}
        L_{push}= -\frac{1}{M}\sum_{m=1}^{M}{\sum_{m'=1}^{M}{ \mathbbm{1}_{[m \neq m']}|| \mu_{m} - \mu_{m'}||_2} } \\
\end{gathered}
\label{equ:ae_push}
\end{equation}

\subsubsection{Total loss}
Equ.~(\ref{equ:total_loss}) shows the total loss we use to train MIX'EM. 
\begin{equation}
\begin{gathered}
        L_{total} = L_{contrast} + \lambda_1 L_{comp-ent} + \lambda_2 L_{inst-ent} + \lambda_3 L_{push} + \lambda_4 L_{pull} 
        \\
\end{gathered}
\label{equ:total_loss}
\end{equation}

\subsection{Clustering to acquire classification}
Once MIX'EM is trained, we apply the K-means algorithm to the representations or the embeddings to generate the final clusters. Our experiments show that K-means on representations delivers superior performance. We also tried using other off-the-shelf clustering methods including spectral clustering \cite{von2007tutorial}, and obtained similar results. Moreover, the dominant mixture component index also provides a highly accurate classification, as shown in the next section.

\section{Experiments}
We experiment with three standard datasets, STL10 \cite{coates2011analysis}, CIFAR10 \cite{krizhevsky2009learning}, and CIFAR100$-$20 \cite{krizhevsky2009learning} (CIFAR100 with 20 super-classes). STL10 is a subset of ImageNet designed to benchmark self-supervised and unsupervised methods. It includes 100k  unlabeled images and train/test labeled splits with 5k/8k images. The labeled splits have ten categories, and the unlabeled split includes a similar but broader set of categories. We use ResNet-18 \cite{he2016deep} for all the experiments. Since CIFAR10 and CIFAR100-20 images are already in small resolution (32x32), we remove the down-sampling in the first convectional layer and the max-pooling layer for experiments on them.

We assume to know the number of categories in the datasets and use the same number of mixture components for the main results, but also provide results with larger number of components. As a potential application in the real world, consider labeling items from a set of already known classes in a supermarket \cite{han2020automatically} or in a warehouse. Nevertheless, if the number of classes is not known in advance, there are techniques to estimate it \cite{han2019learning}.

For each dataset, we first train the bare SimCLR for 600 epochs with embedding dimension 256 without the mixture embedding component. For this stage, on STL10 we use the train and unlabled splits, and on CIFAR10/100-20 we use the train split. We call this model "base SimCLR encoder." Next, we continue training with/without the mixture embedding module with a lower embedding dimension to investigate various aspects under equal conditions. We set embedding dimension to 32, as it gives slightly better results. In the rest of the paper, by "SimCLR", we mean the version trained further on the base SimCLR encoder.



For the evaluation of the semantic clustering, we use the three popular metrics: clustering accuracy (ACC), normalized mutual information (NMI), and adjusted rand index (ARI). For ACC, we use the Hungarian method to map cluster indices to the ground-truth labels. Following the standard practice for unsupervised settings \cite{ji2019invariant,yang2016joint}, we train MIX'EM on the combination of all labeled data and evaluate on the test split. To gain a better understanding of the role of data in unsupervised learning, in ablation studies, we also provide separate evaluations for when test data is not used in the training of MIX'EM. Unless specified otherwise, the results are obtained by taking the average of five separate trainings and are accompanied by the standard deviation. 

Training hyper-parameters, including learning rate, mini-batch size, learning schedule, the loss weight terms ($\lambda_1$, $\lambda_2$, $\lambda_3$, and $\lambda_4$), and augmentation configuration are determined by trying few different values for each dataset. Details for the training setup are provided in the supplementary material. Given enough computational resources, we believe that extending the experiments to a larger dataset like ImageNet would be straightforward.





\setlength{\tabcolsep}{4pt}
\begin{table}
\begin{center}
\caption{Effect of different MIX'EM loss terms when evaluated on STL10. Randomly initialized K-means is repeated $50$ for times. "Max Comp" means doing clustering by using the dominant component indices as the final cluster labels.}
\label{table:incremental}
\resizebox{1.\textwidth}{!}{
\begin{tabular}{ll|cc|cc|cc}
\hline
 & &  \multicolumn{2}{c|}{\bf \thead{ACC}} & \multicolumn{2}{c|}{\bf NMI} & \multicolumn{2}{c}{\bf ARI} \\ 
 & \bf Method & \multicolumn{1}{c}{\bf \thead{K$-$means}} & \multicolumn{1}{c|}{\bf Max Comp}& \multicolumn{1}{c}{\bf \thead{K$-$means}} & \multicolumn{1}{c|}{\bf Max Comp}& \multicolumn{1}{c}{\bf \thead{K$-$means}} & \multicolumn{1}{c}{\bf Max Comp} \\ \hline
 
(1) & SimCLR & $65.01 \pm 1.79$ & $-$ & $66.2 \pm 0.66$ & $-$ & $43.94 \pm 1.09$ & $-$ \\ \hline
(2) & $+$ representation normalization & $69.95 \pm 0.78$ & $-$ & $67.05\pm 0.33$  & $-$ & $55.37 \pm 0.78$ & $-$ \\ \hline
 &  \multicolumn{7}{l}{ MIX'EM} \\ 
(3) & $+ L_{comp-ent}$ & $69.88 \pm 0.17$ & $31.83 \pm 0.29$ & $66.52 \pm 0.08$ & $20.61 \pm 0.15$ &  $54.43 \pm 0.23$ & $12.15 \pm 0.17$ \\ \hline
(4) & $+ L_{inst-ent}$ & $76.21 \pm 0.12$ & $73.45 \pm 0.03$ & $67.27 \pm 0.07$ & $64.3 \pm 0.04$ &  $60.16 \pm 0.16$ & $55.88 \pm 0.05$ \\ \hline
(5) & $+$ MIX'EM initializes K$-$means & $76.21 \pm 0.09$ & $73.45 \pm 0.03$ & $67.23 \pm 0.08$ & $64.3 \pm 0.04$ & $60.12 \pm 0.11$ & $55.88 \pm 0.05$  \\ \hline
(6) & $+ L_{push} + L_{pull}$ & $\textbf{77.76} \pm \textbf{0.08}$ &  $68.44 \pm 0.44$ & $\textbf{68.03} \pm \textbf{0.07}$ & $64.08 \pm 0.1$ & $\textbf{61.35} \pm \textbf{0.1}$ & $54.66 \pm 0.09$ \\ \hline
(7) & $-$ MIX'EM initializes K$-$means & $70.78 \pm 0.19$ & $68.44 \pm 0.44$ & $67.57 \pm 0.42$ & $64.08 \pm 0.1$ & $55.95 \pm 0.16$ & $54.66 \pm 0.09$ \\
\hline
\end{tabular} 
}

\end{center}
\end{table}
\setlength{\tabcolsep}{1.4pt}

\begin{figure}[t]
\centering
\includegraphics[width=110mm]{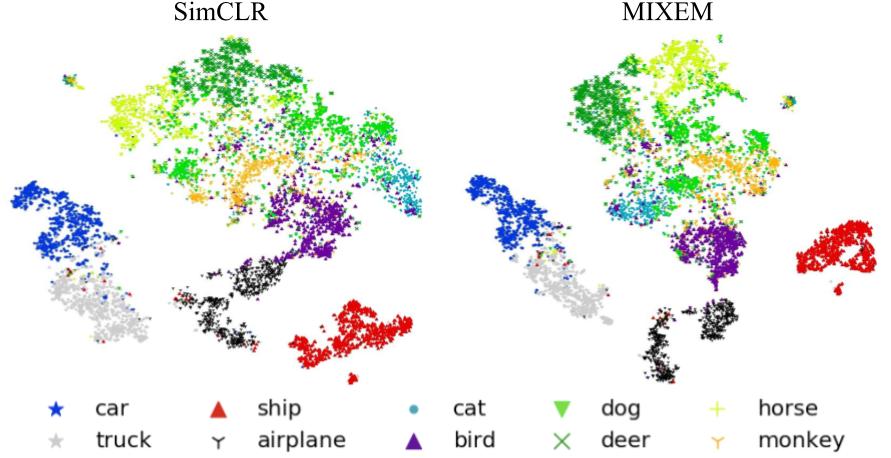} 
\caption{tSNE visualization of representation on STL10 to illustrate the category-level disentanglement. Samples are marked and colored based on their true label.
}
\label{fig:tsne_rep_fc}
\end{figure}

\subsection{Results}
We begin by analyzing the effect of each loss term on STL10. Table \ref{table:incremental} shows that, starting from contrastive loss alone (SimCLR) \cite{chen2020simple}, gradually adding various MIX'EM loss terms consistently improves the performance. Row (2) illustrates the importance of normalizing the representations before applying K-means. Rows (4) vs. (5), and (6) vs. (7) show using MIX'EM to initialize the K-means results in significant improvement.

In figure \ref{fig:tsne_rep_fc} we present the tSNE \cite{maaten2008visualizing} visualization of the representations for SimCLR and MIX'EM. In line with the quantitative evaluation, MIX'EM representations are more disentangled and constitute more compact clusters. Using contrastive loss alone does not adequately pull samples from similar categories to each other, resulting in a sparse space. The mixture module guides the training process via mixing coefficients, forcing the encoder to allocate more compact regions to different categories.

\begin{figure}[t]
\centering
\includegraphics[width=9cm]{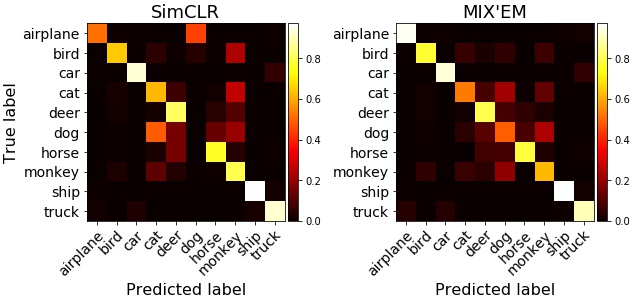} 
\caption{Confusion matrix for prediction vs. true label on STL10. }
\label{fig:confusion}
\end{figure}

In figure \ref{fig:tsne_fcmax}, the top row displays the tSNE visualization of the embeddings for the dominant component of each image as we gradually add MIX'EM loss terms. The bottom row shows how, in turn, the mixture embeddings get more disentangled as we do so. For a category level analysis of MIX'EM, we show the accuracy confusion matrix in figure \ref{fig:confusion}. The animal categories are clearly more difficult to discriminate and benefit the most from MIX'EM.  In the supplementary material, we provide more visualizations that indicate how the correct category is very hard to recognize in some images.

\subsection{Comparison to the state-of-the-art}
Table \ref{table:sota} compares performance of MIX'EM to the state-of-the-art. On the more challenging dataset of STL10, our model outperforms all other works by a large margin. On CIFAR, our model outperforms other works, except SCAN \cite{van2020learning} (concurrent work) when further trained with a classification objective. MIX'EM has very low standard deviation, which would be of high importance in a real-world application. On STL10 and CIFAR100-20, standard deviation of SCAN is about an order of magnitude higher. Since MIX'EM improves representations in terms of separability, SCAN should benefit from using MIX'EM as the representation encoder. On CIFAR100-20 for all models, the results are generally worse compared to other datasets. This is mainly due to the some confusing mapping of classes to super-classes. For example, "bicycle" and "train" both are mapped to "vehicles 1" and most animals are divided based on size, rather than semantics.

\setlength{\tabcolsep}{4pt}
\begin{table}[t]
\begin{center}
\caption{Comparison to the state-of-the-art in unsupervised image classification.}
\label{table:sota}
\resizebox{1.\textwidth}{!}{
\begin{threeparttable}
\begin{tabular}{c|ccc|ccc|ccc}
\hline
\multicolumn{1}{c}{\bf \thead{Dataset}} &\multicolumn{3}{|c}{\bf \thead{CIFAR10}} &\multicolumn{3}{|c}{\bf \thead{CIFAR100-20}} &\multicolumn{3}{|c}{\bf \thead{STL10}}
\\\hline
\multicolumn{1}{c}{\bf \thead{Metric}} & \multicolumn{1}{|c}{\bf \thead{ACC}} & \multicolumn{1}{c}{\bf \thead{NMI}} & \multicolumn{1}{c}{ \bf \thead{ARI}}& \multicolumn{1}{|c}{\bf \thead{ACC}} & \multicolumn{1}{c}{\bf \thead{NMI}} & \multicolumn{1}{c}{\bf \thead{ARI}}& \multicolumn{1}{|c}{\bf \thead{ACC}} & \multicolumn{1}{c}{\bf \thead{NMI}} & \multicolumn{1}{c}{\bf \thead{ARI}} \\ \hline

\thead{Linear classifier on SimCLR\\
(supervised)} & $89.6 \pm 0.2$ & $79.91 \pm 0.3$ & $79.15 \pm 0.35$ & $79.69 \pm 0.15$ &  $64.38 \pm 0.15$ & $61.54 \pm 0.26$ & $87.22 \pm 0.09$ & $77.1 \pm 0.13$ & $74.88\pm 0.15$ \\ \hline
DEC \cite{xie2016unsupervised} & $30.1$ & $25.7$ & $16.1$ & $18.5$ & $13.6$ & $5.0$ & $35.9$ & $27.6$ & $18.6$ \\ 
DAC \cite{chang2017deep} & $52.2$ & $40.0$ & $30.1$ & $23.8$ & $18.5$ & $8.8$ & $47.0$ & $36.6$ & $25.6$ \\ 
DeepCluster \cite{caron2018deep} & $37.4$ & $-$ & $-$ & $18.9$ & $-$ & $-$ & $65.6$ & $-$ & $-$ \\ 
ADC \cite{haeusser2018associative} & $32.5$ & $-$ & $-$ & $16.0$ & $-$ & $-$ & $53$ & $-$ & $-$ \\ 
PICA \cite{huang2020deep} & $0.561$ & $0.645$ & $0.467$ & $-$ & $-$ & $-$ & $0.592$ & $0.693$ & $0.504$ \\
IIC \cite{ji2019invariant} & $61.7$ & $51.1$ & $41.1$ & $25.7$ & $22.5$ & $11.7$ & $59.6$ & $49.6$ & $39.7$ \\
SCAN \cite{van2020learning} & $81.8 \pm 0.3$ & $71.2 \pm 0.4$ & $66.5 \pm 0.4$ & $42.2 \pm 3.0$ & $44.1 \pm 1.0$ & $26.7 \pm 1.3$ & $75.5 \pm 2.0$ & $65.4 \pm 1.2$ & $59.0 \pm 1.6$ \\ 
SCAN  \cite{van2020learning} w/ classification & $\textbf{87.6} \pm 0.4$ & $\textbf{78.7} \pm 0.5$ & $\textbf{75.8} \pm 0.7$ & $\textbf{45.9} \pm 2.7$ & $\textbf{46.8} \pm 1.3$ & $\textbf{30.1} \pm 2.1$ & $76.7 \pm 1.9$ & $\textbf{68.0} \pm 1.2$ & \textbf{$\textbf{61.6} \pm 1.8$ }\\ 
SimCLR + K-means & $79.72 \pm 0.22$ & $69.56 \pm 0.28$ & $62.06 \pm 0.43$ & $42.58 \pm 0.74$ & $43.7 \pm 0.59$ & $24.43 \pm 0.81$ &  $69.95 \pm 0.78$ & $67.05 \pm 0.33$ & $55.37 \pm 0.78$ \\ \hline
MIX'EM + K-means &  $81.87 \pm 0.23$ & $70.85 \pm 0.26$ & $66.59 \pm 0.37$ & $43.77 \pm 0.51$ & $\textbf{46.41} \pm 0.11$ & $27.12 \pm 0.33$ & $\textbf{77.76} \pm 0.08$ & $\textbf{68.03} \pm 0.07$ & $\textbf{61.35} \pm 0.1$\\
MIX'EM max component& $82.19 \pm 0.21$ & $71.35 \pm 0.27$ & $67.15 \pm 0.32$ & $39.19 \pm 0.44$ & $43.59 \pm 0.14$ & $26.67 \pm 0.12$ & $68.44 \pm 0.44$ & $64.08 \pm 0.1$ & $54.66 \pm 0.09$ \\ \hline
\end{tabular}
\end{threeparttable}
}
\end{center}
\end{table}
\setlength{\tabcolsep}{1.4pt}

\subsection{Ablation studies}

\subsubsection{Number of the mixture components}
Although we set the number of mixture components to be the same as the number of categories, it is not necessary to do so. With 20 and 40 components on STL10, clustering accuracy is relatively stable: $76.22\%$ and $75.53\%$, respectively, compared to $77.76\%$ with 10 components. In these cases, where we have more mixture components than classes, we initialize K-Means using the most frequent components. As MIX’EM is a solution for clustering to a known number of categories, we believe it is optimal to use that information in the design.
\subsubsection{Initializing K-means using MIX'EM}
\label{sec:init_kmena}
\begin{figure}[t]
\centering

\subfloat[SimCLR]{\label{fig:kmean_random_simclr}\includegraphics[height=2cm]{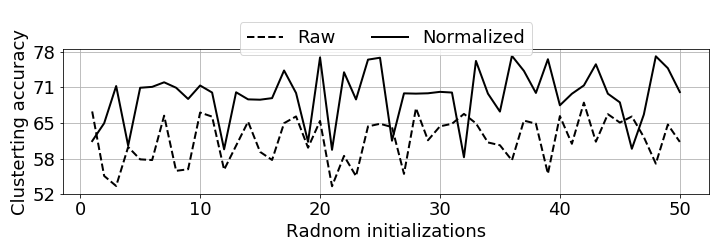} }
\subfloat[MIX'EM]{\label{fig:kmean_random_mixem}\includegraphics[height=2cm]{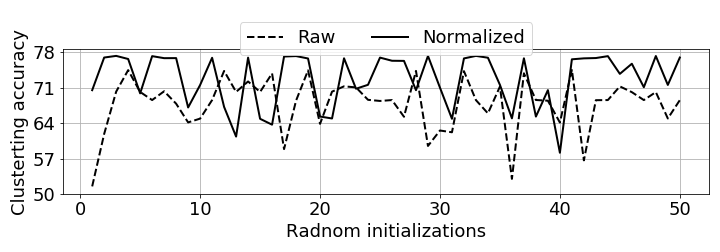}}

\caption{K-means with standard random initialization fluctuates heavily over  different runs and does not guarantee converging to an optimal solution (on STL10). 
}
\label{fig:kmean_random}
\end{figure}

K-means lacks a robust initialization method and the standard procedure is to run K-means many times using random initialization and choose the best one in terms of inertia. We experimented with up to 100 runs and found 50 times to work the best on our models. However, this is neither reliable nor efficient on large scale datasets. With random initialization, K-means is not guaranteed to find the best clustering within practical limits (see large fluctuations in accuracy across different runs in figure \ref{fig:kmean_random}). Running K-means for 50 times on representations of dimensionality 512 takes about 21 seconds on the relatively small STL10 test split (8k images and 10 classes). On 10k images of CIFAR100-20/CIFAR100 with 20/100 classes it takes 112/221 seconds on average. This will get worse on larger datasets with even more categories. 



In MIX'EM, we use the mean of representations by each component, based on samples with the same dominant mixture component, to initialize K-means. This eliminates need for multiple random initializations, while consistently delivering higher accuracy. Rows (4),(5),(6) and (7) in table \ref{table:incremental} show the performance with MIX'EM initialization. In particular, rows (6) and (7) illustrate how K-means with 50 random initialization can be far worse than using MIX'EM for initialization. A single run with MIX'EM initialization, on average, takes $0.27$ , $1.5$, and $3$ seconds on STL10, CIFAR100-20, and CIFAR100, in order.

\subsubsection{Effect on contrastive representation learning}
In MIX'EM, the contrastive loss is vital for successful training. This raises the question of how the mixture module influences the performance of representations in terms of accuracy of a linear classifier trained on the frozen features generated by the encoder, which is the standard measure to evaluate self-supervised representations \cite{zhang2016colorful,oord2018representation,chen2020simple}. To answer this, we train a linear classifier on the frozen base SimCLR encoder, SimCLR, and various forms of MIX'EM. According to table~\ref{table:linear}, the mixture module neither improves nor hurts the representation quality the linear classification task. This implies that the representation learned using SimCLR contain rich information enough to easily train a supervised linear classifier without further disentanglement of the representation. However, for the unsupervised setting, category-level disentanglement of the representation seems essential, as we observed a significant boost in clustering accuracy using MIX'EM.

\setlength{\tabcolsep}{4pt}
\begin{table}
\begin{center}
\caption{MIX'EM does not disrupt contrastive learning objective while imposing category-level disentanglement on representations. (evaluated on STL10)}
\label{table:linear}
\resizebox{.8\textwidth}{!}{
\begin{tabular}{llc}
 & \bf Model & \bf \thead{Supervised linear  classifier accuracy} \\ \hline
 (0) & base SimCLR encoder & $86.1$\\ \hline
(1) & SimCLR & $87.21 \pm 0.1$\\ \hline
 & MIX'EM \\ 
(2) &+ entropy maximization &  $87.25 \pm 0.05$  \\ \hline
(3) & + component entropy minimization &  $87.15 \pm 0.06$ \\ \hline
(4) & + associative embedding loss & $87.22 \pm 0.09$  \\ \hline
\end{tabular}
}
\end{center}
\end{table}
\setlength{\tabcolsep}{1.4pt}

\subsubsection{Effect of using test data in training}
We investigate three scenarios regarding data splits used for training of MIX'EM and SimCLR; (1) using both train and test splits for training. This is the standard setting as we do not use the available labels for training \cite{ji2019invariant,yang2016joint}. (2) only using the train split for training; (3) using the train and unlabeled splits (on STL10 only) for training. Note that we always evaluate on the test split. The results are presented in table \ref{table:train_data_effect}.

\setlength{\tabcolsep}{4pt}
\begin{table}
\begin{center}
\caption{The effect of data splits used for training MIX'EM and SimCLR on the K-Menas clustering performance. All evaluations are on test split.}
\label{table:train_data_effect}
\resizebox{.8\textwidth}{!}{
\begin{tabular}{l|cc|l|l|l}
\hline
 \bf Dataset & \bf  \thead{Training splits} & \bf Method & \bf \thead{ACC} & \bf \thead{NMI} & \bf\thead{ARI} \\ \hline
 
 \multirow{6}{*}{\bf \thead{STL10}} & \multirow{2}{*}{\bf \thead{train+unlabled}} & SimCLR & $67.44 \pm 0.71$ & $64.90 \pm 0.1$ & $51.26 \pm 0.18$ \\ 

 & & MIX'EM & $71.04 \pm 1.13$ & $62.56 \pm 0.85$ & $52.29 \pm 1.41$  \\ \cline{2-6}
 
 & \multirow{2}{*}{\bf \thead{train}} & SimCLR & $65.57 \pm 0.4$ & $63.72 \pm 0.2$ & $50.50 \pm 0.38$ \\
 & & MIX'EM & $74.20 \pm 0.06$ & $65.19 \pm 0.06$ & $55.89 \pm 0.1$ \\ \cline{2-6}
 
 & \multirow{2}{*}{\bf \thead{train+test}} & SimCLR & $69.95 \pm 0.78$ & $67.05 \pm 0.33$ & $55.37 \pm 0.78$ \\ 
 
 & & MIX'EM & $77.76 \pm 0.08$ & $68.03 \pm 0.07$ & $61.35 \pm 0.1$ \\ \hline \hline
 \multirow{4}{*}{\bf \thead{CIFAR10}} & \multirow{2}{*}{\bf \thead{train}} & SimCLR & $77.74 \pm 0.08$ & $67.21 \pm 0.15$ & $58.54 \pm 0.16$ \\ 
 & & MIX'EM & $79.51 \pm 0.41$ & $68.29 \pm 0.28$ & $63.29 \pm 0.44$ \\ \cline{2-6}
 
 & \multirow{2}{*}{\bf \thead{train+test}} & SimCLR & $79.72 \pm 0.22$ & $69.56 \pm 0.28$ & $62.06 \pm 0.43$ \\ 
 & & MIX'EM & $81.87 \pm 0.23$ & $70.85 \pm 0.26$ & $66.59 \pm 0.37$ \\ \hline 

\end{tabular} 
}
\end{center}
\end{table}
\setlength{\tabcolsep}{1.4pt}
\subsubsection{Scenario (1) vs (2)}
Using test split in training consistently improves performance, having a more significant impact on STL10. We argue that this is due to the size and visual difficulty of STL10. CIFAR10 has 50k training and 10k test images. But, on STL10 there is only 5k training and 8k test images. Hence, on STL10, using test split in training means 160\% additional data, while on CIFAR10 it is just a 20\% addition. In the future, a more controlled experiment by progressively removing fractions of training data should be helpful for making a more informed conclusion. Additionally, STL10 is a subset of ImageNet and is visually more complex. On CIFAR100-20 trend is quite similar to CIFAR10.
\subsubsection{Scenario (2) vs. (3)} Unlabeled split of STL10 contains 100k images; however, we do not know the distribution of the categories, and it contains unknown distractor categories. Therefore, despite increasing training data by a large factor, performance drops in this scenario. MIX'EM presumes access to the expected number of categories, which does not hold for the unlabeled set. We believe this is the reason why the accuracy of K-means on SimCLR does not drop as much in this case. Nevertheless, MIX'EM still is significantly more accurate.

\section{Conclusion}

We presented MIX'EM, a novel solution for unsupervised image classification. MIX'EM builds a mixture of embeddings module into SimCLR in order to impose semantic structure on the representations. To successfully train MIX'EM, we introduce various loss terms. MXI'EM sets a new stat-of-the-art unsupervised accuracy on STL10 and performs on par with current models on CIFAR. We also show that applying K-means itself on normalized representations from SimCLR results in impressively high accuracy. We believe this can be used as a new measure for evaluating the quality of self-supervised representation learning methods. The results we publish here could be further improved by using the latest findings in contrastive visual representation learning \cite{tian2020makes}. In the future, we would like to explore the impact of our model on image retrieval and instance segmentation tasks. Moreover, studying the theoretical aspects of MIX'EM could provide insight for further improvements. 

\section*{Acknowledgments}
This work was partially funded by the FWO SBO project HAPPY.

\bibliographystyle{splncs}
\bibliography{accv2020cameraready}

\begin{thebibliography}{10}

\bibitem{he2016deep}
He, K., Zhang, X., Ren, S., Sun, J.:
\newblock Deep residual learning for image recognition.
\newblock In: Proceedings of the IEEE conference on computer vision and pattern
  recognition. (2016)  770--778

\bibitem{krizhevsky2012imagenet}
Krizhevsky, A., Sutskever, I., Hinton, G.E.:
\newblock Imagenet classification with deep convolutional neural networks.
\newblock In: Advances in neural information processing systems. (2012)
  1097--1105

\bibitem{sun2017revisiting}
Sun, C., Shrivastava, A., Singh, S., Gupta, A.:
\newblock Revisiting unreasonable effectiveness of data in deep learning era.
\newblock In: Proceedings of the IEEE international conference on computer
  vision. (2017)  843--852

\bibitem{he2019rethinking}
He, K., Girshick, R., Doll{\'a}r, P.:
\newblock Rethinking imagenet pre-training.
\newblock In: Proceedings of the IEEE International Conference on Computer
  Vision. (2019)  4918--4927

\bibitem{lin2014microsoft}
Lin, T.Y., Maire, M., Belongie, S., Hays, J., Perona, P., Ramanan, D.,
  Doll{\'a}r, P., Zitnick, C.L.:
\newblock Microsoft coco: Common objects in context.
\newblock In: European conference on computer vision, Springer (2014)  740--755

\bibitem{deng2009imagenet}
Deng, J., Dong, W., Socher, R., Li, L.J., Li, K., Fei-Fei, L.:
\newblock Imagenet: A large-scale hierarchical image database.
\newblock In: 2009 IEEE conference on computer vision and pattern recognition,
  Ieee (2009)  248--255

\bibitem{bachman2019learning}
Bachman, P., Hjelm, R.D., Buchwalter, W.:
\newblock Learning representations by maximizing mutual information across
  views.
\newblock In: Advances in Neural Information Processing Systems. (2019)
  15535--15545

\bibitem{gidaris2018unsupervised}
Gidaris, S., Singh, P., Komodakis, N.:
\newblock Unsupervised representation learning by predicting image rotations.
\newblock arXiv preprint arXiv:1803.07728 (2018)

\bibitem{wu2018unsupervised}
Wu, Z., Xiong, Y., Yu, S.X., Lin, D.:
\newblock Unsupervised feature learning via non-parametric instance
  discrimination.
\newblock In: Proceedings of the IEEE Conference on Computer Vision and Pattern
  Recognition. (2018)  3733--3742

\bibitem{oord2018representation}
Oord, A.v.d., Li, Y., Vinyals, O.:
\newblock Representation learning with contrastive predictive coding.
\newblock arXiv preprint arXiv:1807.03748 (2018)

\bibitem{chen2020simple}
Chen, T., Kornblith, S., Norouzi, M., Hinton, G.:
\newblock A simple framework for contrastive learning of visual
  representations.
\newblock arXiv preprint arXiv:2002.05709 (2020)

\bibitem{he2020momentum}
He, K., Fan, H., Wu, Y., Xie, S., Girshick, R.:
\newblock Momentum contrast for unsupervised visual representation learning.
\newblock In: Proceedings of the IEEE/CVF Conference on Computer Vision and
  Pattern Recognition. (2020)  9729--9738

\bibitem{tian2019contrastive}
Tian, Y., Krishnan, D., Isola, P.:
\newblock Contrastive multiview coding.
\newblock arXiv preprint arXiv:1906.05849 (2019)

\bibitem{asano2019critical}
Asano, Y.M., Rupprecht, C., Vedaldi, A.:
\newblock A critical analysis of self-supervision, or what we can learn from a
  single image.
\newblock arXiv preprint arXiv:1904.13132 (2019)

\bibitem{mclachlan1988mixture}
McLachlan, G.J., Basford, K.E.:
\newblock Mixture models: Inference and applications to clustering. Volume~84.
\newblock M. Dekker New York (1988)

\bibitem{jordan1994hierarchical}
Jordan, M.I., Jacobs, R.A.:
\newblock Hierarchical mixtures of experts and the em algorithm.
\newblock Neural computation \textbf{6} (1994)  181--214

\bibitem{bishop1994mixture}
Bishop, C.M.:
\newblock Mixture density networks.
\newblock (1994)

\bibitem{greff2019multi}
Greff, K., Kaufman, R.L., Kabra, R., Watters, N., Burgess, C., Zoran, D.,
  Matthey, L., Botvinick, M., Lerchner, A.:
\newblock Multi-object representation learning with iterative variational
  inference.
\newblock arXiv preprint arXiv:1903.00450 (2019)

\bibitem{chen2019unsupervised}
Chen, M., Arti{\`e}res, T., Denoyer, L.:
\newblock Unsupervised object segmentation by redrawing.
\newblock In: Advances in Neural Information Processing Systems. (2019)
  12726--12737

\bibitem{li2019generating}
Li, C., Lee, G.H.:
\newblock Generating multiple hypotheses for 3d human pose estimation with
  mixture density network.
\newblock In: Proceedings of the IEEE Conference on Computer Vision and Pattern
  Recognition. (2019)  9887--9895

\bibitem{lee2015m}
Lee, S., Purushwalkam, S., Cogswell, M., Crandall, D., Batra, D.:
\newblock Why m heads are better than one: Training a diverse ensemble of deep
  networks.
\newblock arXiv preprint arXiv:1511.06314 (2015)

\bibitem{ye2018occlusion}
Ye, Q., Kim, T.K.:
\newblock Occlusion-aware hand pose estimation using hierarchical mixture
  density network.
\newblock In: Proceedings of the European Conference on Computer Vision (ECCV).
  (2018)  801--817

\bibitem{makansi2019overcoming}
Makansi, O., Ilg, E., Cicek, O., Brox, T.:
\newblock Overcoming limitations of mixture density networks: A sampling and
  fitting framework for multimodal future prediction.
\newblock In: Proceedings of the IEEE Conference on Computer Vision and Pattern
  Recognition. (2019)  7144--7153

\bibitem{varamesh2020mixture}
Varamesh, A., Tuytelaars, T.:
\newblock Mixture dense regression for object detection and human pose
  estimation.
\newblock In: Proceedings of the IEEE/CVF Conference on Computer Vision and
  Pattern Recognition. (2020)  13086--13095

\bibitem{newell2017pixels}
Newell, A., Deng, J.:
\newblock Pixels to graphs by associative embedding.
\newblock In: Advances in neural information processing systems. (2017)
  2171--2180

\bibitem{newell2017associative}
Newell, A., Huang, Z., Deng, J.:
\newblock Associative embedding: End-to-end learning for joint detection and
  grouping.
\newblock In: Advances in Neural Information Processing Systems. (2017)
  2277--2287

\bibitem{chen2020improved}
Chen, X., Fan, H., Girshick, R., He, K.:
\newblock Improved baselines with momentum contrastive learning.
\newblock arXiv preprint arXiv:2003.04297 (2020)

\bibitem{gutmann2010noise}
Gutmann, M., Hyv{\"a}rinen, A.:
\newblock Noise-contrastive estimation: A new estimation principle for
  unnormalized statistical models.
\newblock In: Proceedings of the Thirteenth International Conference on
  Artificial Intelligence and Statistics. (2010)  297--304

\bibitem{hjelm2018learning}
Hjelm, R.D., Fedorov, A., Lavoie-Marchildon, S., Grewal, K., Bachman, P.,
  Trischler, A., Bengio, Y.:
\newblock Learning deep representations by mutual information estimation and
  maximization.
\newblock arXiv preprint arXiv:1808.06670 (2018)

\bibitem{donahue2016adversarial}
Donahue, J., Kr{\"a}henb{\"u}hl, P., Darrell, T.:
\newblock Adversarial feature learning.
\newblock arXiv preprint arXiv:1605.09782 (2016)

\bibitem{dumoulin2016adversarially}
Dumoulin, V., Belghazi, I., Poole, B., Mastropietro, O., Lamb, A., Arjovsky,
  M., Courville, A.:
\newblock Adversarially learned inference.
\newblock arXiv preprint arXiv:1606.00704 (2016)

\bibitem{doersch2015unsupervised}
Doersch, C., Gupta, A., Efros, A.A.:
\newblock Unsupervised visual representation learning by context prediction.
\newblock In: Proceedings of the IEEE international conference on computer
  vision. (2015)  1422--1430

\bibitem{caron2018deep}
Caron, M., Bojanowski, P., Joulin, A., Douze, M.:
\newblock Deep clustering for unsupervised learning of visual features.
\newblock In: Proceedings of the European Conference on Computer Vision (ECCV).
  (2018)  132--149

\bibitem{noroozi2016unsupervised}
Noroozi, M., Favaro, P.:
\newblock Unsupervised learning of visual representations by solving jigsaw
  puzzles.
\newblock In: European Conference on Computer Vision, Springer (2016)  69--84

\bibitem{zhang2016colorful}
Zhang, R., Isola, P., Efros, A.A.:
\newblock Colorful image colorization.
\newblock In: European conference on computer vision, Springer (2016)  649--666

\bibitem{van2020learning}
Van~Gansbeke, W., Vandenhende, S., Georgoulis, S., Proesmans, M., Van~Gool, L.:
\newblock Learning to classify images without labels.
\newblock arXiv preprint arXiv:2005.12320 (2020)

\bibitem{huang2020deep}
Huang, J., Gong, S., Zhu, X.:
\newblock Deep semantic clustering by partition confidence maximisation.
\newblock In: Proceedings of the IEEE/CVF Conference on Computer Vision and
  Pattern Recognition. (2020)  8849--8858

\bibitem{ji2019invariant}
Ji, X., Henriques, J.F., Vedaldi, A.:
\newblock Invariant information clustering for unsupervised image
  classification and segmentation.
\newblock In: Proceedings of the IEEE International Conference on Computer
  Vision. (2019)  9865--9874

\bibitem{asano2019self}
Asano, Y.M., Rupprecht, C., Vedaldi, A.:
\newblock Self-labelling via simultaneous clustering and representation
  learning.
\newblock arXiv preprint arXiv:1911.05371 (2019)

\bibitem{yang2016joint}
Yang, J., Parikh, D., Batra, D.:
\newblock Joint unsupervised learning of deep representations and image
  clusters.
\newblock In: Proceedings of the IEEE Conference on Computer Vision and Pattern
  Recognition. (2016)  5147--5156

\bibitem{yan2020clusterfit}
Yan, X., Misra, I., Gupta, A., Ghadiyaram, D., Mahajan, D.:
\newblock Clusterfit: Improving generalization of visual representations.
\newblock In: Proceedings of the IEEE/CVF Conference on Computer Vision and
  Pattern Recognition. (2020)  6509--6518

\bibitem{zhan2020online}
Zhan, X., Xie, J., Liu, Z., Ong, Y.S., Loy, C.C.:
\newblock Online deep clustering for unsupervised representation learning.
\newblock In: Proceedings of the IEEE/CVF Conference on Computer Vision and
  Pattern Recognition. (2020)  6688--6697

\bibitem{sanchez2019learning}
Sanchez, E.H., Serrurier, M., Ortner, M.:
\newblock Learning disentangled representations via mutual information
  estimation.
\newblock arXiv preprint arXiv:1912.03915 (2019)

\bibitem{von2007tutorial}
Von~Luxburg, U.:
\newblock A tutorial on spectral clustering.
\newblock Statistics and computing \textbf{17} (2007)  395--416

\bibitem{coates2011analysis}
Coates, A., Ng, A., Lee, H.:
\newblock An analysis of single-layer networks in unsupervised feature
  learning.
\newblock In: Proceedings of the fourteenth international conference on
  artificial intelligence and statistics. (2011)  215--223

\bibitem{krizhevsky2009learning}
Krizhevsky, A., Hinton, G.,  et~al.:
\newblock Learning multiple layers of features from tiny images.
\newblock (2009)

\bibitem{han2020automatically}
Han, K., Rebuffi, S.A., Ehrhardt, S., Vedaldi, A., Zisserman, A.:
\newblock Automatically discovering and learning new visual categories with
  ranking statistics.
\newblock arXiv preprint arXiv:2002.05714 (2020)

\bibitem{han2019learning}
Han, K., Vedaldi, A., Zisserman, A.:
\newblock Learning to discover novel visual categories via deep transfer
  clustering.
\newblock In: Proceedings of the IEEE International Conference on Computer
  Vision. (2019)  8401--8409

\bibitem{maaten2008visualizing}
Maaten, L.v.d., Hinton, G.:
\newblock Visualizing data using t-sne.
\newblock Journal of machine learning research \textbf{9} (2008)  2579--2605

\bibitem{xie2016unsupervised}
Xie, J., Girshick, R., Farhadi, A.:
\newblock Unsupervised deep embedding for clustering analysis.
\newblock In: International conference on machine learning. (2016)  478--487

\bibitem{chang2017deep}
Chang, J., Wang, L., Meng, G., Xiang, S., Pan, C.:
\newblock Deep adaptive image clustering.
\newblock In: Proceedings of the IEEE international conference on computer
  vision. (2017)  5879--5887

\bibitem{haeusser2018associative}
Haeusser, P., Plapp, J., Golkov, V., Aljalbout, E., Cremers, D.:
\newblock Associative deep clustering: Training a classification network with
  no labels.
\newblock In: German Conference on Pattern Recognition, Springer (2018)  18--32

\bibitem{tian2020makes}
Tian, Y., Sun, C., Poole, B., Krishnan, D., Schmid, C., Isola, P.:
\newblock What makes for good views for contrastive learning.
\newblock arXiv preprint arXiv:2005.10243 (2020)

\end{thebibliography}

\end{document}